\documentclass[10pt,journal,letter]{IEEEtran}
\usepackage{fancyhdr, epsfig, epsf, amsthm, amsmath, amssymb, amsfonts, color}
\usepackage{enumerate}
\usepackage{dsfont}
\usepackage[noadjust]{cite}

\usepackage{stfloats}
\usepackage{algorithm}
\usepackage{algpseudocode}

\usepackage{rotating,booktabs,multirow}

\usepackage{graphicx,subcaption}
\usepackage{enumitem}

\newtheorem{proposition}{Proposition}

\def\({\left(}
\def\){\right)}

\def\[{\left[}
\def\]{\right]}

\def\sm{\small}
\def\nm{\normalsize}

\usepackage{array}

\newcommand{\chapternote}[1]{%
 \let\thempfn\relax
  \footnotetext[0]{\emph{#1}}
  }

\begin{document}

\title{\LARGE  Mix2FLD: Downlink Federated Learning After \\Uplink Federated Distillation With Two-Way Mixup}

\author{ Seungeun Oh, \IEEEauthorrefmark{2}Jihong Park, Eunjeong Jeong, \IEEEauthorrefmark{3}Hyesung Kim, \IEEEauthorrefmark{4}Mehdi Bennis, and Seong-Lyun Kim   
	      \vskip -13pt 
	\thanks{This work was partly supported by Institute of Information \& communications Technology Planning \& Evaluation (IITP) grant funded by the Korea government (MSIT) (No.2018-0-00170, Virtual Presence in Moving Objects through 5G), Basic Science Research Program through the National Research Foundation of Korea(NRF) funded by the Ministry of Science and ICT(NRF-2017R1A2A2A05069810), the Academy of Finland Project MISSION, SMARTER, and the 2019 EU-CHISTERA Projects LeadingEdge and CONNECT.}\vskip -20pt
	      
	\thanks{S. Oh, E. Jeong, and S.-L. Kim are with the School of Electrical and Electronic Engineering, Yonsei University, 120-749 Seoul, Korea (email: \{seoh, ejjeong, slkim\}@ramo.yonsei.ac.kr).}  
	\thanks{\IEEEauthorrefmark{2}J. Park is with the School of Information Technology, Deakin University, Geelong, VIC 3220, Australia (email: jihong.park@deakin.edu.au).}
\thanks{\IEEEauthorrefmark{3}H. Kim is with Samsung Research, Samsung Electronics, Seoul, Korea (email: hye1207@gmail.com).}
\thanks{\IEEEauthorrefmark{4}M. Bennis is with the Centre for Wireless Communications, University of Oulu, 90500 Oulu, Finland (email: mehdi.bennis@oulu.fi).}

}
\vskip -15pt
\maketitle
\vskip -17pt
\begin{abstract}
This letter proposes a novel communication-efficient and privacy-preserving distributed machine learning framework, coined \emph{Mix2FLD}. To address uplink-downlink capacity asymmetry, local model outputs are uploaded to a server in the uplink as in federated distillation (FD), whereas global model parameters are downloaded in the downlink as in federated learning (FL). This requires a model output-to-parameter conversion at the server, after collecting additional data samples from devices. To preserve privacy while not compromising accuracy, linearly mixed-up local samples are uploaded, and inversely mixed up across different devices at the server. Numerical evaluations show that Mix2FLD achieves up to $16.7$\% higher test accuracy while reducing convergence time by up to $18.8$\% under asymmetric uplink-downlink channels compared to FL.
\end{abstract}
\vspace{-5pt}

\vspace{-15pt}
\maketitle

\section{Introduction}
User-generated local data is essential in training machine learning (ML) models for mission-critical applications, but exchanging data may violate privacy and induce huge communication overhead~\cite{Park:2018aa}. Federated learning (FL) is a compelling solution that collectively trains on-device ML models using their local private data~\cite{mm,Yang:FLSurvey}. FL preserves data privacy, in a way that devices only upload their local model parameters to a server over wireless links, and download their average global model parameters. However, the communication efficiency of FL is problematic in deep neural network models (DNNs), since its payload sizes increase with the model sizes. The problem is aggravated in the uplink whose channel capacity is more limited by lower transmission power and bandwidth than downlink channels, i.e., uplink-downlink asymmetric channels \cite{JHParkTWC:15}. Federated distillation (FD) resolves this problem, by exchanging model outputs~\cite{Jeong2018,Cha2019,Ahn:2019aa,Park:2018ab}. Regardless of model sizes (e.g., millions of parameters in DNNs), communication payload sizes of FD are fixed as the model output dimension (e.g., 10 labels in MNIST), although FD compromises accuracy.

\begin{figure}
  \begin{subfigure}{\columnwidth}\centering
  \includegraphics[width= .8\columnwidth]{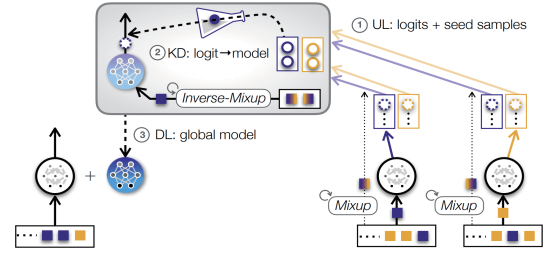} \vskip -2pt
	\caption{\footnotesize\textbf{Mix2FLD}: downlink federated learning (FL) \& uplink federated distillation~(FD) with two-way Mixup (Mix2up) seed sample collection.}
  \end{subfigure}\vskip -8pt
  \vspace{10pt}

	\begin{subfigure}{\columnwidth}\centering
  \includegraphics[width= .85\columnwidth]{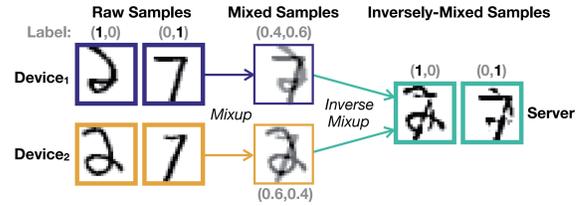} \vskip -5.5pt
  \caption{\footnotesize \textbf{Mix2up}: mixing raw samples at devices \& inversely mixing them across different devices at the server (mixing ratio {$\lambda=0.4$}).}
  \end{subfigure}
  \vskip -8pt
  \caption{\small An illustration of (a) Mix2FLD operation and (b) Mix2up.}
  \vskip -24pt
\end{figure}
		In order to achieve both high accuracy and communication-efficiency under uplink-downlink asymmetric channels, we propose a distributed ML framework, dubbed \emph{Mix2FLD}. As depicted in Fig.~1, Mix2FLD is built upon two key algorithms: \emph{federated learning after distillation (FLD)}~\cite{Park:2018ab} and \emph{Mixup} data augmentation~\cite{Zhang2018}.
		Specifically, by leveraging FLD, each device in Mix2FLD uploads its local model outputs as in FD, and downloads the model parameters as in FL, thereby coping with the uplink-downlink channel asymmetry. Between the uplink and downlink, the server runs knowledge distillation (KD) that transfers a teacher's knowledge (i.e., average outputs, see Sec~\ref{Sec:SysML2}) to an untrained student model (i.e., a global model)~\cite{HintonKD:14}. This output-to-model conversion requires additional training samples collected from devices, which may violate local data privacy while incurring huge communication overhead.
	\vskip -3pt			
		To preserve the data privacy with minimal communication overhead during the seed sample collection, Mix2FLD utilizes a novel \emph{two-way Mixup algorithm (Mix2up)}, as illustrated in Fig.~1b. To hide raw samples, each device in Mix2up uploads locally superpositioned samples using Mixup~\cite{Zhang2018}. Next, before running KD at the server, the uploaded mixed-up samples are superpositioned across different devices, in a way that the resulting sample labels are in the same form of the raw sample labels. This inverse-Mixup provides more realistic synthetic seed samples for KD, without restoring raw samples. Furthermore, with $N_S$ uploaded samples, it can generate $N_I\geq{N_S}$ samples, enabling KD with minimal uplink cost.

		Numerical simulations corroborate that Mix2FLD achieves higher test accuracy and faster training convergence than FL and FD, under both identically and independently distributed (IID) and non-IID local datasets. Furthermore, it is shown that Mix2FLD achieves higher accuracy while preserving more data privacy, compared to FLD using only Mixup (MixFLD), highlighting the importance of Mix2up.

  \vskip -10pt

\begin{table}[t!]\footnotesize
  \centering
  \caption{List of Notations.}\vskip -5pt
    \resizebox{\columnwidth}{!}{\begin{tabular} {l l}
    \toprule[1pt]
    \textbf{Notation} & \textbf{Meaning} \\
    \midrule
    $\mathcal{D}$ & a set of devices\\

    $\mathcal{D}^p$ & a set of uploading success devices at the $p$-th global update\\

    $N_{ch}$ & \# of uplink channels\\
    $N_\textsf{mod}$ & \# of model weights\\

    $N_L$ & \# of ground-truth labels\\

	$(N_S, N_I)$ & (\# of mixed-up samples, \# of inversely mixed-up samples) \\

    $\mathbf{S}_d$ & local dataset of the $d$-th device\\

    $\mathbf{s}_d^{[i]}$ & $i$-th sample in $\mathbf{S}_d$\\

    \;\;{\tiny$\bullet$\nm} $X^{[i]}$ & unlabeled sample of $\mathbf{s}_d^{[i]}=\{X^{[i]},\mathbf{L}_n^{[i]}\}$\\

    \;\;{\tiny$\bullet$\nm} $\mathbf{L}_n^{[i]}$ & label vector of $\mathbf{s}_d^{[i]}$ with the $n$-th label as the ground-truth\\

    $\hat{\mathbf{s}}_d^{[i,j]}$ & mixed-up sample, from $\mathbf{s}_d^{[i]}$ and $\mathbf{s}_d^{[j]}$\\

    $\tilde{\mathbf{s}}_{d,d',n}^{[i,j][i'\!,j']}$ & inversely mixed-up sample, from $\hat{\mathbf{s}}_d^{[i,j]}$\! and $\hat{\mathbf{s}}_{d'}^{[i',j']}$\\

    $\mathbf{G}^p_{\textsf{mod}}$ & global weight vector at the $p$-th global update\\

    \;\;{\tiny$\bullet$\nm} $\mathbf{w}_{d}^{(k)}$ & local weight vector at the $k$-th iteration\\

    $\mathbf{G}^p_{\textsf{out},n}$ & global output vector with the $n$-th ground-truth label\\

    \;\;{\tiny$\bullet$\nm} $\mathbf{F}_{d,n}^{[i_{k}]}$ & local output vector (i.e., softmax logits) of the sample $i_k$\\

    \;\;{\tiny$\bullet$\nm} $\mathbf{\bar{F}}^p_{d,n}$ & local average output vector with the $n$-th ground-truth label\\

    $K$ & \# of local iterations per global update\\

    $K_{s}$ & \# of output-to-parameter converting iterations per global update\\


    $T^y$ & \# of time slots for uploading or downloading $B^y$ bits\\


    \bottomrule[1pt]

  \end{tabular}}
  \vskip -10pt   
\end{table}

\section{System Model}

This section describes our baseline distributed ML architecture and operations, followed by communication channel models. The network under study comprises a set {\sm$\mathcal{D}$\nm} of devices connected to a single server through wireless links. Following a data-parallel distributed ML architecture~\cite{Park:2018aa}, every device owns its local private dataset and an on-device ML model having $N_\textsf{mod}$ weights. The $d$-th device has a local dataset {\sm$\mathbf{S}_d$\nm} of samples, in which the $i$-th sample {\sm$\mathbf{s}_d^{[i]} = \{X^{[i]},\mathbf{L}_n^{[i]}\}$\nm} comprises a pair of an unlabeled sample {\sm$X^{[i]}$\nm} and its label vector {\sm$\mathbf{L}_n^{[i]}=\{\ell_1^{[i]},\ell_2^{[i]},\cdots, \ell_{N_L}^{[i]}\!\}$\nm}. The label vector's element $\ell_n^{[i]}$ equals $1$ if the $n$-th label is the ground-truth, and $0$ otherwise. 

With {\sm$\mathbf{S}_d$\nm}, each device collaboratively trains its model so as to predict the labels of unlabeled data samples in a multi-class classification task. These distributed ML operations are divided into local updates at devices and global updates at the server, as detailed next under FL and FD.  \vskip -15pt   
     \vskip -15pt    \subsection{FL}     \label{Sec:SysML}
   \vskip -2pt   
For each global update, every device updates its local model weights by running $K$ iterations of the stochastic gradient descent algorithm (SGD). At the $k$-th local iteration, the $d$-th device randomly selects the $i_k$-th sample, and updates its \emph{local weight vector} $\mathbf{w}_d^{(k)}$ with a constant learning rate $\eta$~as:

\vspace{-10pt}\sm\begin{align}
\mathbf{w}_d^{(k+1)} = \mathbf{w}_d^{(k)} - \eta \nabla \phi(\mathbf{F}_{d,n}^{[i_{k}]}, \mathbf{L}_n^{[i_k]} | \mathbf{w}_d^{(k)}),\label{Eq:weightFL}
\end{align}\nm
by calculating the gradient of a cross-entropy loss function {\sm$\phi(\mathbf{F}_{d,n}^{[i_{k}]}, \mathbf{L}_n^{[i_k]} | \mathbf{w}_d^{(k)})=-\sum_{m=1}^{N_L} \ell_m^{[i_k]}\log F_{m}^{[i_k]}$\nm}. The term {\sm$\mathbf{F}_{d,n}^{[i_{k}]}$\nm} is the \emph{local output vector} {\sm$\mathbf{F}_{d,n}^{[i_{k}]}=\{F_{1}^{[i_k]},F_{2}^{[i_k]},\cdots,F_{N_L}^{[i_k]} \}$\nm} implying the prediction distribution over $N_L$ labels when the $n$-th label is the ground-truth. The elements of {\sm$\mathbf{F}_{d,n}^{[i_{k}]}$\nm} are softmax normalized logits at the model's last layer, satisfying {\sm$\sum_{m=1}^{N_L} F_{m}^{[i_k]}=1$ with $F_{m}^{[i_k]}\in[0,1]$\nm} {\sm$\forall m$\nm.

After $K$ local iterations, following~\cite{mm}, the $d$-th device in FL uploads its \emph{latest weight vector} {\sm$\mathbf{w}_d^{(K)}$\nm} to the server over a wireless link. A set {\sm$\mathcal{D}^{p}$\nm} of devices can successfully upload the weight vectors at the $p$-th global update, depending on the channel conditions that will be elaborated in Sec.~\ref{Sec:SysWirelress}. 
By taking a weighted average proportional to the number of samples $|\mathcal{D}^p|$ devices have, the server produces the \emph{global weight vector} {\sm$\mathbf{G}_\textsf{mod}^{p}=\sum_{d\in{\mathcal{D}^{p}}}{\vert\mathbf{S}_d\vert\mathbf{w}_d^{(K)}}/\sum_{d\in{\mathcal{D}^{p}}}{\vert\mathbf{S}_d}\vert$\nm} that is downloaded by each device.
 Finally, the $d$-th device replaces {\sm$\mathbf{w}_d^{(K)}$ with $\mathbf{G}_\textsf{mod}^{p}$\nm}, and continues its local updates in \eqref{Eq:weightFL} until the $(p+1)$-th global update. These operations are iterated until {\sm$|\mathbf{G}_{\textsf{mod}}^{p}\!-\!\mathbf{G}_{\textsf{mod}}^{p-1}|/|\mathbf{G}_{\textsf{mod}}^{p-1}|<\varepsilon$\nm} is satisfied, for a constant $\varepsilon>0$.  \vskip -10pt

 \vskip -5pt \subsection{FD} \label{Sec:SysML2}
   \vskip -4pt   
At the $p$-th global update, following~\cite{Jeong2018}, the $d$-th device in FD uploads $N_L$ \emph{local average output vectors}, produced by averaging the local output vectors {\sm$\{\mathbf{F}_{d,n}^{[i_{k}]}\}$\nm} during $K$ local SGD iterations, separately for each ground-truth label. For the $n$-th ground-truth label (i.e., {\sm$\ell_n^{[i_k]}=1$\nm}), the local average output vector {\sm$\bar{\mathbf{F}}_{d,n}^{p}$\nm} is given~as:

\vskip -12pt 
  \sm\begin{align}
    \bar{\mathbf{F}}_{d,n}^{p}=\sum_{k=1}^{K}{\mathds{1}\(\ell_n^{[i_k]}=1\) \mathbf{F}_{d}^{[i_{k}]}}/\sum_{k=1}^K \mathds{1}\(\ell_n^{[i_k]}=1\), \label{eq:FD}
    \end{align}\nm
    \vskip -6pt 
where $\mathds{1}(A)$ becomes $1$ if $A$ is true, and $0$ otherwise. By averaging {\sm$\{\bar{\mathbf{F}}_{d,n}^{p}\}$\nm} across $|\mathcal{D}^p|$ devices, the server generates the \emph{global average output vector} {\sm$\mathbf{G}_{\textsf{out},n}^{p}=\sum_{d\in{\mathcal{D}^{p}}} \bar{\mathbf{F}}_{d,n}^{p}/|\mathcal{D}^p|=\{G_1,G_2,\cdots,G_{N_L}\}$\nm} that is downloaded by each device.

Next, until the $(p+1)$-th global update, the $d$-th device updates its local weight vector $\mathbf{w}_d^{(k)}$ using SGD with KD as:

\vskip-10pt
\sm\begin{align}
\hspace{-5pt}\mathbf{w}_d^{(k+1)} \!=\! \mathbf{w}_d^{(k)} \!-\! \eta \nabla\! \( \phi(\mathbf{F}_{d,n}^{[i_{k}]}, \mathbf{L}_n^{[i_k]} | \mathbf{w}_d^{(k)}) \!+\! \beta \psi(\mathbf{F}_{d,n}^{[i_k]} , \mathbf{G}_{\textsf{out},n}^p) \!\)\!, \label{eq:FDKD}
\end{align}\nm
\noindent with a constant $\beta>0$. In contrast to \eqref{Eq:weightFL}, this includes a \emph{distillation regularizer} {\sm$\psi(\mathbf{F}_{d,n}^{[i_k]} , \mathbf{G}_{\textsf{out},n}^p)= \sum_{m=1}^{N_L} G_{m}\log F_{m}^{[i_k]}$\nm} that measures the gap between $\mathbf{F}_{d,n}^{[i_k]}$ and $\mathbf{G}_{\textsf{out},n}^p$ using cross-entropy. If this knowledge gap is negligible, the device's weight is updated based on its own prediction, and otherwise perturbed proportionally to the gap. These operations continue until {\sm$|\mathbf{G}_{\textsf{out},n}^{p}\!-\mathbf{G}_{\textsf{out},n}^{p-1}|/|\mathbf{G}_{\textsf{out},n}^{p-1}|<\varepsilon$\nm} is satisfied for all $n$.  \vskip -10pt

\vskip -4pt \subsection{Wireless Channel Model} \label{Sec:SysWirelress}

At each global update, we consider uplink unicast and downlink multicast transmissions. 
In the uplink, 
the server allocates equal bandwidth $W^\text{up} = W N_{ch} / |\mathcal{D}|$ to each device for frequency division multiple access (FDMA), whereas in the downlink it utilizes the entire bandwidth $W^\text{dn} = W$. Let the superscript $y=\{\text{up},\text{dn}\}$ identify uplink and downlink. With the transmission power $P^y$ and the distance $r_d$ from the $d$-th device to the server, the received signal-to-noise ratio (SNR) in either uplink or downlink at the $t$-th time slot is $\textsf{SNR}_{d,t}^y = P^y h_{d,t} {r_d}^{-\alpha}/(W^y N_0)$, where $N_0$ is the noise power spectral density, and $\alpha\geq 2$ denotes the path loss exponent. Following Rayleigh block fading channels, the term $h_{d,t}$ is an exponential random variable with unitary mean, independent and identically distributed (IID) across different devices and time slots.

For a target SNR $\theta^y>0$, each received signal is successfully decoded if {\sm$\textsf{SNR}_{d,t}^y \geq \theta^y$\nm}. During $T$ time slots, the received {\sm$B_\text{RX}^y$\nm} bits is thereby given as:

\vspace{-12pt}\sm\begin{align}
B_\text{RX}^y(T) = \tau \sum_{t=1}^{T} \mathds{1}(\textsf{SNR}_{d,t}^y\geq \theta^y) W^y \log_2(1 + \theta^y), \label{Eq:UL}
\end{align}\nm
\vskip -5pt 
where $\tau$ is the channel coherence time identically set as the unit time slot. The latency {\sm$T^y$\nm} slots (or {\sm$\tau T^y$\nm} seconds) for
 uploading or downloading {\sm$B^y$\nm} bits is the minimum $T$ that satisfies {\sm$B_\text{RX}^y(T) \geq B^y$\nm}. In order to avoid unbounded latency, the server allocates up to $T_\text{max}$ time slots equally to the uplink and downlink. A latency outage occurs when $T^y>T_\text{max}$, incurring a straggling device.

In FL, the payload is the model weights, resulting in {\sm$B^\text{up}=B^\text{dn}=b_{\textsf{mod}} N_\textsf{mod} $\nm} bits, where {\sm$b_{\textsf{mod}}$\nm} is each weight size determined by its arithmetic precision. In FD, {\sm$N_L$\nm} output vectors are exchanged each of which has {\sm$N_L$\nm} elements, leading to {\sm$B^\text{up}=B^\text{dn}=b_{\textsf{out}} {N_L}^{\!2} $\nm} bits, where {\sm$b_{\textsf{out}}$\nm} denotes each output size. \vskip -11pt 

\begin{algorithm}[t]{
		\sm
		\caption{FLD with Mix2up (\textbf{Mix2FLD})}
		\begin{algorithmic}[1]
			\State \textbf{Require:} $\{\mathbf{S}_d\}$ with $d\in \mathcal{D}$, $\lambda\in(0,1)$
			\While  {$|\mathbf{G}_{\textsf{out},n}^{p}\!-\mathbf{G}_{\textsf{out},n}^{p-1}|/|\mathbf{G}_{\textsf{out},n}^{p-1}| \geq \varepsilon$}
			
			
			\State \hspace{-15pt}Device $d\in\mathcal{D}$:  \emph{\hfill$\triangleright$ Output upload}
			
			\State \hspace{-15pt}\quad\textbf{if} $p\!=\!1$ \textbf{generates} $\{\hat{\mathbf{s}}_d^{[i,j]}\}$ via \eqref{Eq:mixup} \textbf{end if} \emph{\hfill$\triangleright$ Mixup}

			\State \hspace{-15pt}\quad\textbf{updates} $\mathbf{w}_d^{(k)}$ in \eqref{Eq:weightFL} and $\bar{\mathbf{F}}_{d,n}^{p}$ in \eqref{eq:FD} for $K$ iterations
			
			\State \hspace{-15pt}\quad\textbf{unicasts} $\{\bar{\mathbf{F}}_{d,n}^{p}\}$ (with $\{\hat{\mathbf{s}}_d^{[i,j]}\}$ \textbf{if} $p=1$) to the server
			

			\State \hspace{-15pt}Server: \emph{\hfill $\triangleright$ Output-to-model conversion}
			
			\State \hspace{-15pt}\quad\textbf{if} {$p\!=\!1$} \textbf{generates} $\{\tilde{\mathbf{s}}_{d,d',n}^{[i,j][i'\!,j']}\}$ via \eqref{eq:invmixup} \textbf{end if} \emph{\hfill $\triangleright$ Inverse-Mixup}
			
			\State \hspace{-15pt}\quad\textbf{computes} $\mathbf{G}_{\textsf{out},n}^p$
			
			\State \hspace{-15pt}\quad\textbf{updates} $\mathbf{w}_s^{(k)}$ via \eqref{eq:FLD} for $K_s$ iterations

			\State \hspace{-15pt}\quad\textbf{broadcasts} $\mathbf{G}_\textsf{mod}^{p}=\mathbf{w}_s^{(K_s)}$ to all devices
			
			\State \hspace{-15pt}$p\gets{p+1}$
			
			
			\State \hspace{-15pt}Device $d\in\mathcal{D}$ \textbf{substitutes} $\mathbf{w}_d^{(0)}$ with $\mathbf{G}_\textsf{mod}^{p}$ \emph{\hfill$\triangleright$ Model download}
			
			\EndWhile    
			
		\end{algorithmic} \nm}     
	
\end{algorithm} 
   
     \section{Mix2FLD: Federated Learning After Distillation With Two-Way Mixup}  
\vspace{-1pt} In this section, we propose the idea of FLD and its two implementations, MixFLD and Mix2FLD. Leveraging the Mixup algorithm~\cite{Zhang2018}, MixFLD enables FLD while preserving data privacy during its seed sample collection. Mix2FLD integrates our novel inverse-Mixup algorithm into MixFLD, further improving accuracy.     
\vspace{-17pt} \subsection{FLD}\vspace{-2pt}
FLD aims to address asymmetric uplink-downlink channel capacity. As shown in Fig.~1a, at the $p$-th global update, the $d$-th device uploads $N_L$ local average output vectors {\sm$\{\mathbf{\bar{F}}_{d,n}^p\}$\nm}, thereby constructing the global average output vector {\sm$\mathbf{G}_{\textsf{out},n}^p$\nm} at the server, as in FD. Then, the device downloads the global weight vector {\sm$\mathbf{G}_{\textsf{mod}}^p$\nm} as in FL. The problem is that the server in FLD lacks {\sm$\mathbf{G}_{\textsf{mod}}^p$\nm}, calling for converting {\sm$\mathbf{G}_{\textsf{out},n}^p$\nm} into {\sm$\mathbf{G}_{\textsf{mod}}^p$\nm}.

\textbf{Output-to-Model Conversion.} The key idea is to transfer the knowledge in {\sm$\mathbf{G}_{\textsf{out},n}^p$\nm} to a global model having the weight vector {\sm$\mathbf{G}_{\textsf{mod}}^p$\nm}. To enable this, at the beginning (i.e., $p=1$), each device uploads $N_s$ seed samples randomly selected from its local dataset. By feeding the collected {\sm$|\mathcal{D}| \cdot N_s $\nm} seed samples, as done in \eqref{eq:FDKD}, the server runs $K_s$ iterations of SGD with KD, thereby updating the global model's weight vector {\sm$\mathbf{w}_s^{(k)}$\nm}~as:

\vskip-10pt
\sm\begin{align}
\hspace{-5pt}\mathbf{w}_s^{(k+1)} \!=\! \mathbf{w}_s^{(k)} \!-\! \eta \nabla\! \( \phi(\mathbf{F}_{s,n}^{[i_{k}]}, \mathbf{L}_n^{[i_k]} | \mathbf{w}_s^{(k)}) \!+\! \beta \psi(\mathbf{F}_{s,n}^{[i_k]} , \mathbf{G}_{\textsf{out},n}^p) \!\)\!, \label{eq:FLD}
\end{align}\nm
\vskip-7pt where {\sm$\mathbf{F}_{s,n}^{[i_{k}]}$\nm} is the global model's output vector if the $n$-th label is the ground-truth. Finally, the server yields {\sm$\mathbf{G}_{\textsf{mod}}^p=\mathbf{w}_s^{(K_s)}$\nm} that is downloaded by every device. The remaining operations follow the same procedure of FL. In FLD, {\sm$B^\text{up}=b_{\textsf{out}} {N_L}^{\!2} + \mathds{1}\{p=1\}b_s N_s $\nm} bits, and {\sm$B^\text{dn}=b_{\textsf{mod}} N_\textsf{mod}$\nm} bits, where $b_s$ is the size of each sample.
  \vskip -12pt   

 \subsection{MixFLD: FLD + Mixup}\vspace{-2pt}
The aforementioned FLD operations include seed sample collection that may violate local data privacy. To mitigate this problem, MixFLD applies the Mixup to the sample collection procedure of FLD, as follows. 

\textbf{Mixup Before Collection.} Before uploading the seed samples, the $d$-th device randomly selects two different raw samples $\mathbf{s}_d^{[i]}$ and $\mathbf{s}_d^{[j]}$ having different labels, i.e., $\mathbf{L}_n^{[i]}\neq \mathbf{L}_m^{[j]}$ with $m\neq n$ and $i\neq j$. With a mixing ratio $\lambda\in(0,0.5)$ given identically for all devices, the device linearly combines these two samples (see Fig.~1b), thereby generating a mixed-up sample $\hat{\mathbf{s}}_d^{[i,j]}$ as:

\vskip -13pt\sm\begin{align}
\hat{\mathbf{s}}_{d}^{[i,j]} = \lambda \mathbf{s}_d^{[i]} + (1-\lambda) \mathbf{s}_d^{[j]}. \label{Eq:mixup}
\end{align}\nm\vskip -4pt
In this way, the device uploads $N_s$ mixed-up samples to the server, and the rest of procedures follow FLD.
\vskip -9pt

  \vspace{-2pt}    \subsection{ Proposed. Mix2FLD: MixFLD + Inverse-Mixup}
   \vskip -1pt
It is observed that MixFLD significantly distorts the seed samples, achieving lower accuracy than FD, in our numerical evaluations in Sec.~IV. To ensure not only data privacy but also high accuracy, we propose Mix2FLD that integrates our novel inverse-Mixup algorithm into MixFLD.

For the sake of explanation, we hereafter focus on a two-device setting, where devices $d$ and $d'$ independently mix up the following two raw samples having symmetric labels.

\begin{itemize}[leftmargin=10pt]
	\vspace{3pt}
  \item \!Device $d$: {\sm$\mathbf{s}_d^{[i]}$\nm} with {\sm$\mathbf{L}_1^{[i]}=\{1,0\}$\nm} and {\sm$\mathbf{s}_d^{[j]}$\nm} with {\sm$\mathbf{L}_2^{[j]}=\{0,1\}$\nm}\vspace{5pt}

\vspace{3pt}
  \item \!Device $d'$\!: {\sm$\mathbf{s}_{d'}^{[i']}$\nm} with {\sm$\mathbf{L}_2^{[i']}=\{0,1\}$\nm} and {\sm$\mathbf{s}_{d'}^{[j']}$\nm} with {\sm$\mathbf{L}_1^{[j']}=\{1,0\}$\nm}
\end{itemize}



\begin{figure*}[t!]
	\centering
	\begin{subfigure}[t]{\columnwidth}
		\includegraphics[width=\columnwidth]{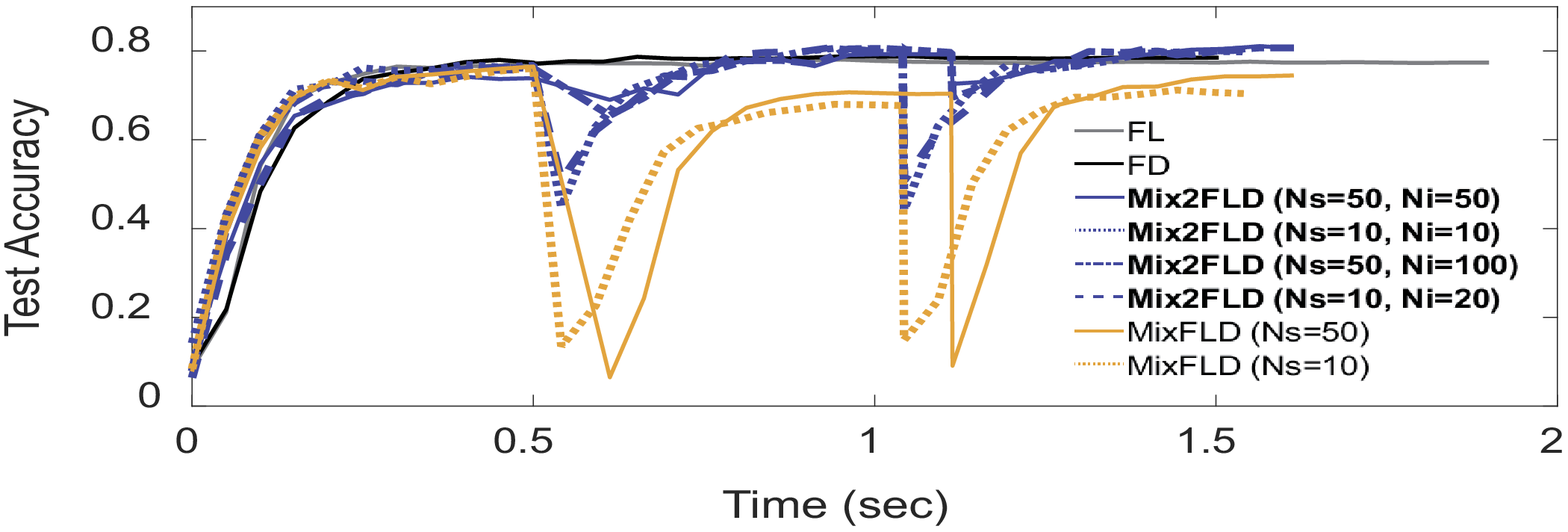} \vskip-5pt
		\caption{Asymmetric channels, IID dataset}
	\end{subfigure}
	\begin{subfigure}[t]{\columnwidth}
		\includegraphics[width=\columnwidth]{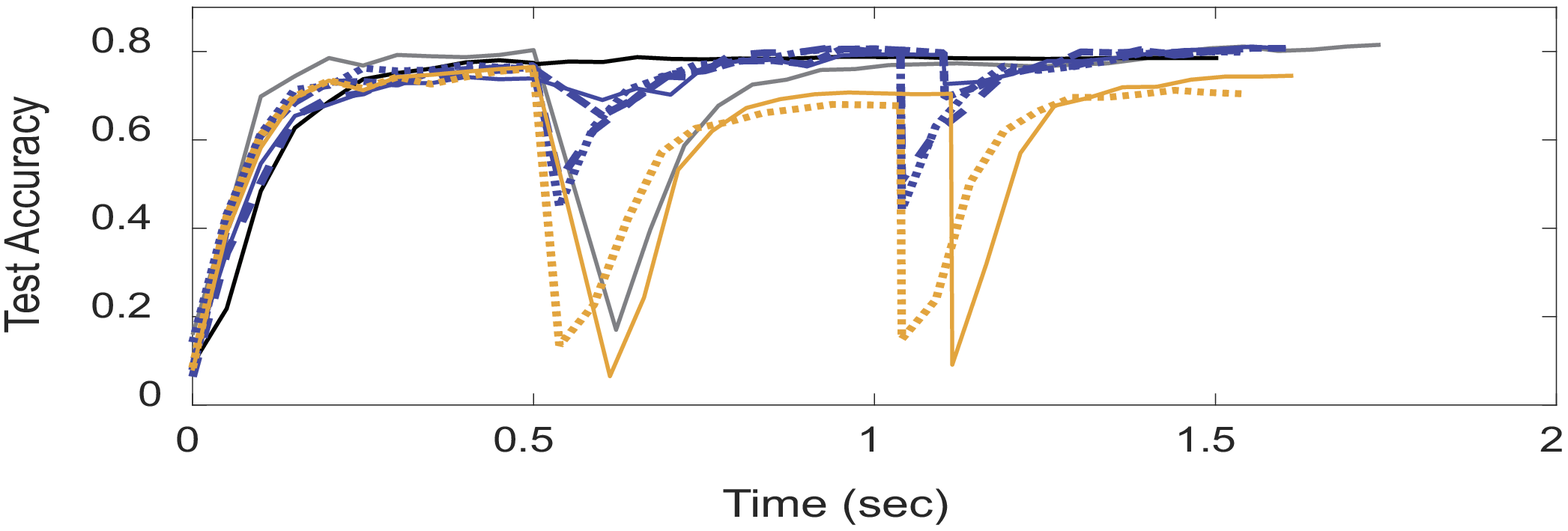} \vskip-5pt
		\caption{Symmetric channels, IID dataset.}
	\end{subfigure}

	\begin{subfigure}[t]{\columnwidth}
		\includegraphics[width=\columnwidth]{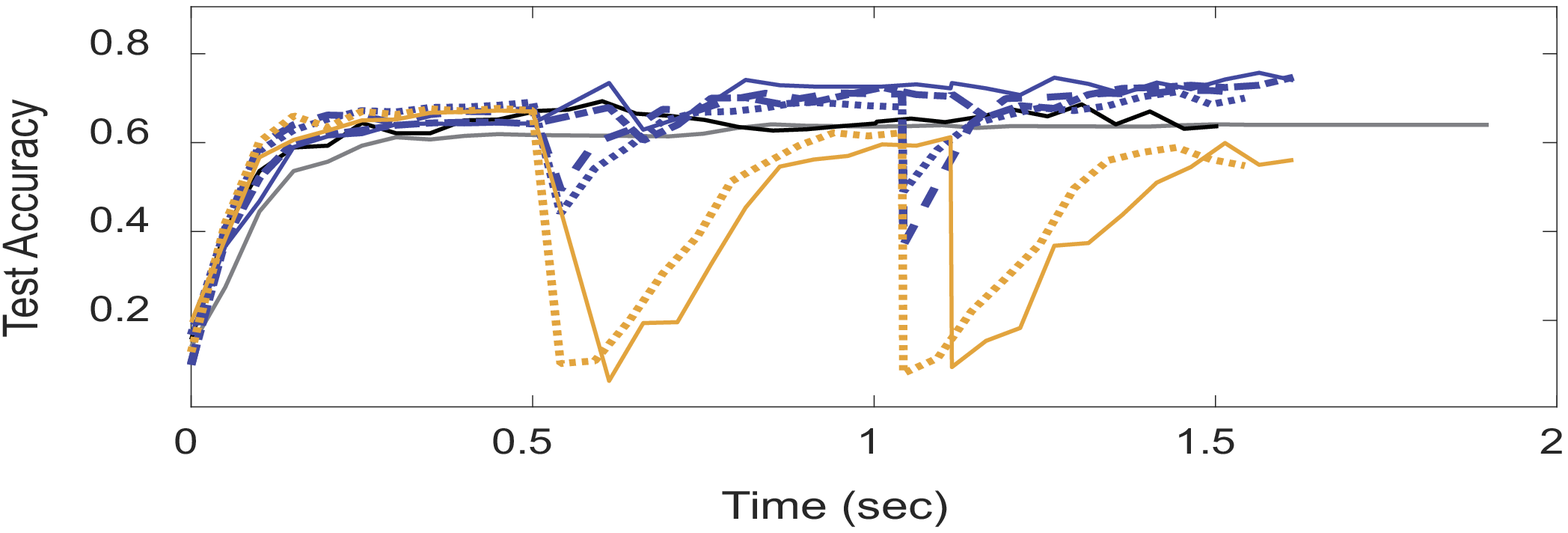} \vskip-5pt
		\caption{Asymmetric channels, Non-IID dataset.}
	\end{subfigure}
	\begin{subfigure}[t]{\columnwidth}
		\includegraphics[width=\columnwidth]{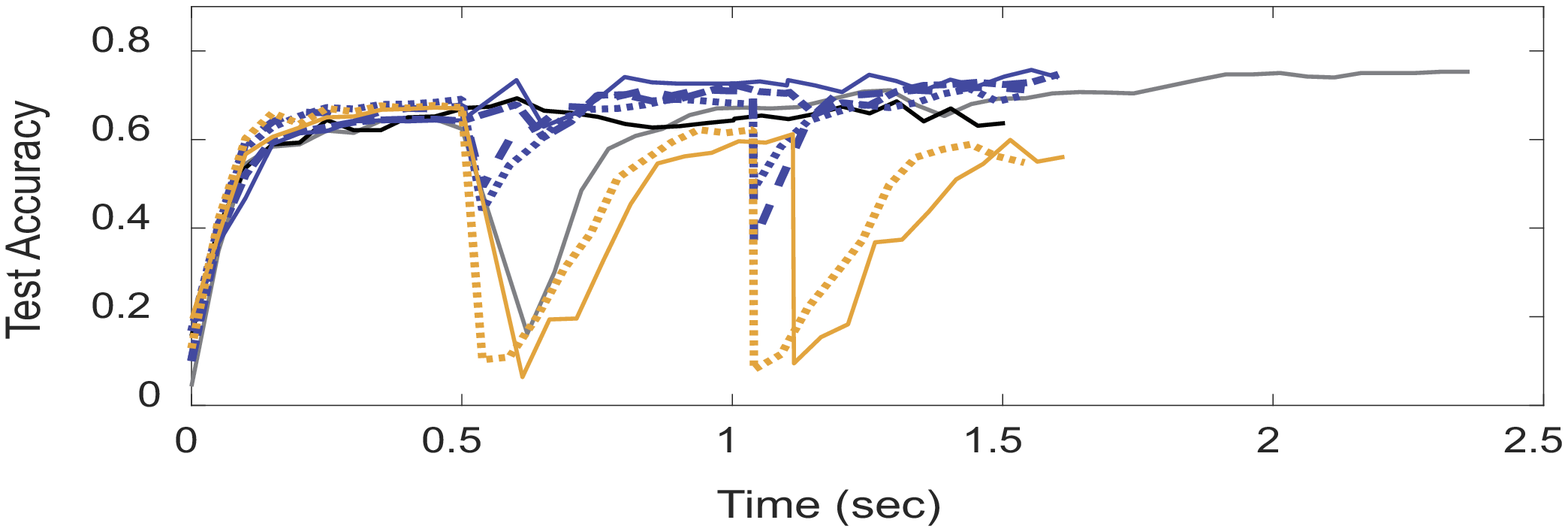} \vskip-5pt
		\caption{Symmetric channels, Non-IID dataset.}
	\end{subfigure}

	\caption{\small Learning curve of a randomly selected device in Mix2FLD, compared to FL, FD, and MixFLD, under asymmetric ($P^\text{up}=23$ dBm, $P^\text{dn}=40$ dBm) and symmetric ($P^\text{up}=P^\text{dn}=40$ dBm) channels, when $\lambda=0.1$ with IID and non-IID datasets.}
	\vskip -15pt   
\end{figure*}

According to \eqref{Eq:mixup}, the mixed-up samples {\sm$\hat{\mathbf{s}}_{d}^{[i,j]}$\nm} and {\sm$\hat{\mathbf{s}}_{d'}^{[i',j']}$\nm} have the \emph{soft labels} $\{\lambda,1-\lambda\}$ and $\{1-\lambda,\lambda\}$, respectively, in contrast to the \emph{hard labels} $\{1,0\}$ and $\{0,1\}$ of raw samples.

\textbf{Inverse-Mixup After Collection.}
Before training the global model using \eqref{eq:FLD}, the sever in Mix2FLD converts the soft labels back into hard labels. To this end, we propose \emph{inverse-Mixup} that linearly combines $N$ mixed-up samples such that the resulting sample has a hard label. 
	For the case of $N=2$, as shown in Fig.~1b, with the above-mentioned symmetric setting, the server combines {\sm$\hat{\mathbf{s}}_d^{[i,j]}$\nm} and {\sm$\hat{\mathbf{s}}_{d'}^{[i',j']}$\nm}, such that the resulting {\sm$\tilde{\mathbf{s}}_{d,d',n}^{[i,j][i'\!,j']}$\nm} has the $n$-th converted hard label as the ground-truth. This is described as:

 \vskip -10pt
  \sm\begin{align}
    \tilde{\mathbf{s}}_{d,d',n}^{[i,j][i'\!,j']}=\hat{\lambda}\hat{\mathbf{s}}_d^{[i,j]}+(1-\hat{\lambda})\hat{\mathbf{s}}_{d'}^{[i',j']}. \label{eq:invmixup}
    \end{align}\nm 
The inverse mixing ratio $\hat{\lambda}$ for $N\geq 2$ is chosen in the following way.
\vspace{-4pt}
   \begin{proposition} 
	When $N$ raw samples are combined with the mixing ratios $(\lambda_1, \lambda_2, \dots, \lambda_N)$, the inverse mixing ratios $(\hat{\lambda}_{1,n},\hat{\lambda}_{2,n},\dots,\hat{\lambda}_{N,n})$ that make an inversely-mixup sample has the $n$-th label as the ground-truth are given by solving the following equation.
  \vskip -12pt   
	\sm\begin{equation}
		\begin{bmatrix}
		\hat{\lambda}_{1,1}       &   \hat{\lambda}_{1,2}       & \dots     &   \hat{\lambda}_{1,N}       \\
		\hat{\lambda}_{2,1}       &   \hat{\lambda}_{2,2}       & \dots     &   \hat{\lambda}_{2,N}       \\
		\vdots  &  \vdots   &   \vdots  &   \vdots  \\   
		\hat{\lambda}_{N,1}       &   \hat{\lambda}_{N,2}       & \dots     &   \hat{\lambda}_{N,N}       \\
		\end{bmatrix}=
		\begin{bmatrix}
		\lambda_1       &   \lambda_2       & \dots     &   \lambda_N       \\
		\lambda_2      &   \lambda_3       & \dots     &   \lambda_1       \\
		\vdots  &  \vdots   &   \vdots  &   \vdots  \\   
		\lambda_N       &   \lambda_1      & \dots     &   \lambda_{N-1}      \\
		\end{bmatrix}^{-1},
	\label{eq:Nsamples}
	\end{equation}\nm  
	  \vspace{-3pt}
	where $\sum_{d=1}^{N}{\lambda_d}=1$. 
	
\end{proposition}

\noindent \emph{Proof:}
First, consider $N=2$.
Suppose the target hard label is $\{1,0\}$, i.e., $n=1$. Applying $\{1,0\}$ to the LHS of \eqref{eq:invmixup} and $\{\lambda,1-\lambda\}$ and $\{1-\lambda,\lambda\}$ of {\sm$\hat{\mathbf{s}}_{d}^{[i,j]}$\nm} and {\sm$\hat{\mathbf{s}}_{d'}^{[i',j']}$\nm} to the RHS of \eqref{eq:invmixup} yields two equations.\sm\begin{align}
1 &= \hat{\lambda} \lambda + (1-\hat{\lambda})(1-\lambda)\\
0 &= \hat{\lambda}(1-\lambda) + (1-\hat{\lambda})\lambda
\end{align}\nm  \vskip -5pt   
Solving these equations yields the desired $\hat{\lambda}$. By induction, this can be generalized to $N\geq 2$, completing the proof. \hfill $\blacksquare$

\noindent 

  Hereafter, for the sake of convenience, we fix $N$ to 2. 
	 By alternating $\hat{\lambda}$ with $n=1$ and $2$, inversely mixing up two mixed-up samples {\sm$\hat{\mathbf{s}}_d^{[i,j]}$\nm} and {\sm$\hat{\mathbf{s}}_{d'}^{[i',j']}$\nm} yields two inversely mixed-up samples {\sm$\tilde{\mathbf{s}}_{d,d',1}^{[i,j][i'\!,j']}$\nm} and {\sm$\tilde{\mathbf{s}}_{d,d',2}^{[i,j][i'\!,j']}$\nm}. The server generates $N_I$ inversely mixed-up samples by pairing two samples with symmetric labels among $N_S$ mixed-up samples. By nature, inverse-Mixup is a data augmentation scheme, so $N_I$ can be larger than $N_S$. Finding the optimal $N_I$ that achieves the highest accuracy with minimal memory usage could be an interesting topic for future work.

Note that none of the raw samples are identical to inversely mixed-up samples. To ensure this, inverse-Mixup is applied only for the seed samples uploaded from different devices, thereby preserving data privacy. The overall operation of Mix2FLD is summarized in Algorithm 1.


  \vspace{-9pt}

\section{Numerical Evaluation and Discussions}

In this section, we numerically evaluate the performance of Mix2FLD compared to FL, FD, and MixFLD, in terms of the test accuracy and convergence time of a randomly selected reference device, under different data distributions (IID and non-IID) and uploaded/generated seed sample configurations ($(N_S, N_I)\in\{(10,10),(10,20),(50,50),(50,100)\}$). The convergence time includes communication delays {\sm$\tau (T^\text{up}+T^\text{dn})$\nm} seconds during the uplink and downlink (see Sec.~\ref{Sec:SysWirelress}), as well as the computing delays of devices and the server, which are measured using tic-toc elapsed time.

Every device has a $3$-layer convolutional neural network model ($2$ convolutional layers, $1$ fully-connected layer) having $N_\textsf{mod}= 12,\!544$. The server's global model follows the same architecture. The model weight and output parameter sizes are given identically as $b_\textsf{mod}=b_\textsf{out}$ = $32$ bits.

Each device owns its local MNIST dataset with $N_L=10$ and $|\mathbf{S}_d|=500$. For the IID case, every label has the same number of samples. For the non-IID case, randomly selected two labels have $2$ samples respectively, while each of the other labels has $62$ samples. Each sample size is given as $b_s = 6,\!272$ bits ($8$ bits $\times$ ($28 \times 28$) pixels).

Other simulation parameters are given as: $|\mathcal{D}|=10$, $K=6,\!400$ iterations, $K_{\text{KD}}=3,\!200$ iterations, $\eta=0.01$, $\varepsilon=0.05$, $\beta$=0.01, $N_{ch}=2$, $W=10$ MHz, $P^\text{up}=23$ dBm, $P^\text{dn}=40$ dBm, $r_d=1$~km, $\alpha=4$, $N_0=-174$ dBm/Hz, $\theta^{\text{up}}=\theta^{\text{dn}}=3$, $\tau=1$ ms, and $T_\text{max}=100$ ms.

\vspace{-2pt}\textbf{Impact of Channel Conditions.}
Fig. 2 shows that Mix2FLD achieves the highest accuracy with moderate convergence under asymmetric channel conditions. Compared to FL uploading model weights, Mix2FLD's model output uploading reduces the uplink payload size by up to $42.4$x. Under asymmetric channels with the limited uplink capacity (Figs. 2a and c), this enables more frequent and successful uploading, thereby achieving up to $16.7$\% higher accuracy and $1.2$x faster convergence. Compared to FD, Mix2FLD leverages the high downlink capacity for downloading the global model weights, which often provides higher accuracy than downloading model outputs as reported in \cite{Jeong2018}. In addition, the global information of Mix2FLD is constructed by collecting seed samples and reflecting the global data distribution, rather than by simply averaging local outputs as used in FD. Thereby, Mix2FLD achieves up to $17.3$\% higher accuracy while taking only $2.5$\% more convergence time than FD. Under symmetric channels, FL achieves the highest accuracy. Nevertheless, Mix2FLD still converges $1.9$x faster than FL, thanks to its smaller uplink payload sizes and more frequent updates. 

\begin{figure}[t!] 
	\begin{center}		
		\includegraphics[width=1.0\linewidth]{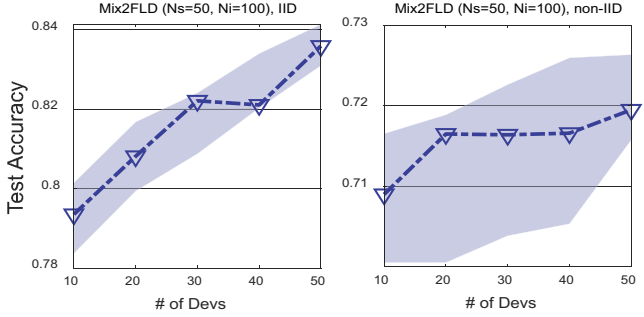}
	\end{center}
	  \vskip -12pt   
	\caption{\small Test accuracy distribution of Mix2FLD w.r.t the number of devices, under symmetric channels with IID and non-IID datasets.}
	  \vskip -12pt   
\end{figure}

\vspace{-2pt} \textbf{Fluctuation of Test Accuracy.} 
FL, MixFLD, and Mix2FLD have instantaneous accuracy drop in global update. After downloading the global information, a noise reflecting global data distributions is inserted into local models, leading to a drastic decrease in test accuracy. This accuracy drop is gradually recovered during local updates, and finally higher accuracy can be achieved than before the noise insertion. In FD, a noise is inserted for every training sample, and partially reflected as an additional loss function, resulting in smaller accuracy drops.

\vskip -2pt\textbf{Impact of the Number of Devices.}
		Fig.~3 shows the scalability of Mix2FLD, under both IID and non-IID data distributions.
		When the number of devices is increased from 10 to 50, the average of test accuracy increases by $5.7$\% and variance decreases by $50$\% with IID dataset. In the non-IID dataset, the test accuracy gain is smaller than that of the IID dataset, while having the same tendency.

  \vskip -2pt   
\textbf{Impact of Mix2up.}
Fig.~2c and Fig.~2d corroborate that Mix2FLD is particularly effective in coping with non-IID data. In our non-IID datasets, samples are unevenly distributed, and locally trained models become more biased, degrading accuracy compared to IID datasets in Fig.~2a and Fig.~2c. This accuracy loss can partly be restored by additional global training (i.e., output-to-model conversion) that reflects the entire dataset distribution using few seed samples. While preserving data privacy, MixFLD attempts to realize this idea. However, as observed in Fig.~2d, MixFLD fails to achieve high accuracy as its mixed-up samples inject too much noise into the global training process. Mix2FLD resolves this problem by utilizing inversely mixed up samples, reducing unnecessary noise. Thanks to its incorporating the data distribution, even under symmetric channels (Fig.~2d), Mix2FLD achieves the accuracy as high as FL.

\vspace{-2pt}\textbf{Latency, Privacy, and Accuracy Tradeoffs.}
For all cases in Fig.~2, in Mix2FLD and MixFLD, reducing the seed sample amount ($N_s=10$) provides faster convergence time in return for compromising accuracy, leading to a \emph{latency-accuracy} tradeoff. Furthermore, in Fig.~2, even if $N_S$ is the same, when $N_I$ is large, the accuracy increases up to $1.7\%$. Such data augmentation effect of inverse-Mixup enables Mix2FLD to effectively increase accuracy without additional latency. Next, we validate the data privacy guarantees of Mixup and Mix2up. This is evaluated using \emph{sample privacy}, given as the minimum similarity between a mixed-up sample and its raw sample: $\log (\min\{||\hat{\mathbf{s}}_d^{[i,j]}\!-\mathbf{s}_d^{[i]}||,\!||\hat{\mathbf{s}}_d^{[i,j]}\!-\mathbf{s}_d^{[j]}||\}\!)$ according to \cite{Mair18,Jeong:FML19}. 
Table II shows that Mixup ($\lambda>0$) with a single device preserves more sample privacy than the case without Mixup ($\lambda=0$). Table III illustrates that Mix2up with two devices preserves higher sample privacy than Mixup thanks to its additional (inversely) mixing up the seed samples across devices. It also shows that each inversely mixed-up sample does not resemble its raw sample but an arbitrary sample having the same ground-truth label. Both Tables II and~III show that the mixing ratio $\lambda$ closer to $0.5$ (i.e., equally mixing up two samples) ensures higher sample privacy, which may require compromising more accuracy. Investigating the \emph{privacy-accuracy} tradeoff is deferred to future work.


  \vskip -9pt


  \vskip -10pt   

\section{Concluding Remarks}  \vskip -3pt   
In this letter, we proposed Mix2FLD that copes with asymmetric uplink-downlink channel capacities, while preserving data privacy. Numerical evaluations corroborated its effectiveness in terms of accuracy and convergence time, under supervised learning in the MNIST classification task. Applying Mix2up to other distributed learning scenarios could be an interesting topic for future research. Also, extending this idea to distributed reinforcement learning by leveraging the proxy experience memory method as in \cite{Cha2019} as well as the convergence analysis of Mix2FLD is left to future work.

  \vskip -9pt   


\begin{table}[t!]\footnotesize
	\centering
	\caption{Sample privacy, \emph{Mixup} ($N_s$=100).}\vskip -5pt
	\resizebox{\columnwidth}{!}{\begin{tabular} {l c c c c c c}
			\toprule[1pt]
			\multirow{2}{*}{\textbf{Dataset}}& \multicolumn{6}{c}{\textbf{Sample Privacy} Under Mixing Ratio $\lambda$} \\
			&  $\lambda$ = 0.001 & 0.1 & 0.2 & 0.3 & 0.4 & 0.499   \\
			\midrule
			MNIST & 2.163 & 4.465 & 5.158 & 5.564 & 5.852 & \textbf{6.055} \\
			FMNIST& 1.825 & 4.127 & 4.821 & 5.226 & 5.514 & \textbf{5.717}  \\
			CIFAR-10& 2.582 & 4.884 & 5.577 & 5.983 & 6.270 & \textbf{6.473}  \\
			CIFAR-100& 2.442 & 4.744 & 5.438 & 5.843 & 6.131 & \textbf{6.334}  \\
			\bottomrule[1pt]
	\end{tabular}}   
	\vskip -5pt   
\end{table} 

\begin{table}[t!]\footnotesize
	\centering
	\caption{ Sample privacy, \emph{Mix2up} ($N_s$=100).}\vskip -5pt
	\resizebox{\columnwidth}{!}{\begin{tabular} {l c c c c c c}
			\toprule[1pt]
			\multirow{2}{*}{\textbf{Dataset}}& \multicolumn{6}{c}{\textbf{Sample Privacy} Under Mixing Ratio $\lambda$} \\
			&  $\lambda$ = 0.001 & 0.1 & 0.2 & 0.3 & 0.4 & 0.499   \\
			\midrule
			MNIST & 2.557 & 4.639 & 5.469 & 6.140 & 7.007 & \textbf{9.366} \\
			FMNIST& 2.196 & 4.568 & 5.410 & 6.143 & 6.925 & \textbf{9.273}  \\
			CIFAR-10& 2.824 & 5.228 & 6.076 & 6.766 & 7.662 & \textbf{10.143}  \\
			CIFAR-100& 2.737 & 5.151 & 6.050 & 6.782 & 7.652 & \textbf{10.104}  \\
			\bottomrule[1pt]
	\end{tabular}}   
	\vskip -10pt
\end{table}
  \vskip -3pt   
\bibliographystyle{ieeetr}

\end{document}